\documentclass[letterpaper, 10 pt, conference]{ieeeconf}
\IEEEoverridecommandlockouts
\usepackage[pdftex]{graphicx}
\usepackage{xcolor}
\usepackage{multirow}
\usepackage{flushend}
\usepackage{mwe}
\usepackage{caption}
\usepackage{subcaption}
\usepackage[utf8]{inputenc}
\usepackage[T1]{fontenc}
\usepackage{pgfplots}
\usepackage{pgfplotstable}
\usepgfplotslibrary{colorbrewer}
\usepgfplotslibrary{statistics}

\colorlet{negro}{black}
\colorlet{gris}{black!70}
\colorlet{rojo}{red!70!black}
\colorlet{rojol}{red}

\title{\LARGE \bf
FastCycle: A Message Sharing Framework for Modular Automated Driving Systems}
\author{Mehdi Testouri$^1$, Gamal Elghazaly$^1$, Raphaël Frank$^1$
\thanks{$^1$Mehdi Testouri, Gamal Elghazaly and  Raphaël Frank are with 360Lab, Interdisciplinary Centre for Security, Reliability and Trust (SnT), University of Luxembourg, L-1855 Luxembourg {\tt\small \{mehdi.testouri, gamal.elghazaly, raphael.frank\}@uni.lu}} 
\thanks{\newline $^2$Paper code is available at: https://gitlab.uni.lu/360lab-public/fastcycle}
}

\begin{document}

\maketitle
\thispagestyle{empty}
\pagestyle{empty}

\begin{abstract}

Automated Driving Systems (ADS) have rapidly evolved in recent years and their architecture becomes sophisticated. Ensuring robustness, reliability and safety of performance is particularly important. The main challenge in building an ADS is the ability to meet certain stringent performance requirements in terms of both making safe operational decisions and finishing processing in real-time. Middlewares play a crucial role to handle these requirements in ADS. The way middlewares share data between the different system components has a direct impact on the overall performance, particularly the latency overhead. To this end, this paper presents \textit{FastCycle} as a lightweight multi-threaded zero-copy messaging broker to meet the requirements of a high fidelity ADS in terms of modularity, real-time performance and security. We discuss the architecture and the main features of the proposed framework. Evaluation of the proposed framework based on standard metrics in comparison with popular middlewares used in robotics and automated driving shows the improved performance of our framework. The implementation of FastCycle and the associated comparisons with other frameworks are open sourced$^2$. 
\end{abstract}

\section{Introduction}
\begin{figure*}[!t]
\centering
\includegraphics[scale = 0.58, clip, trim=1.5cm 5cm 1.5cm 2.55cm]{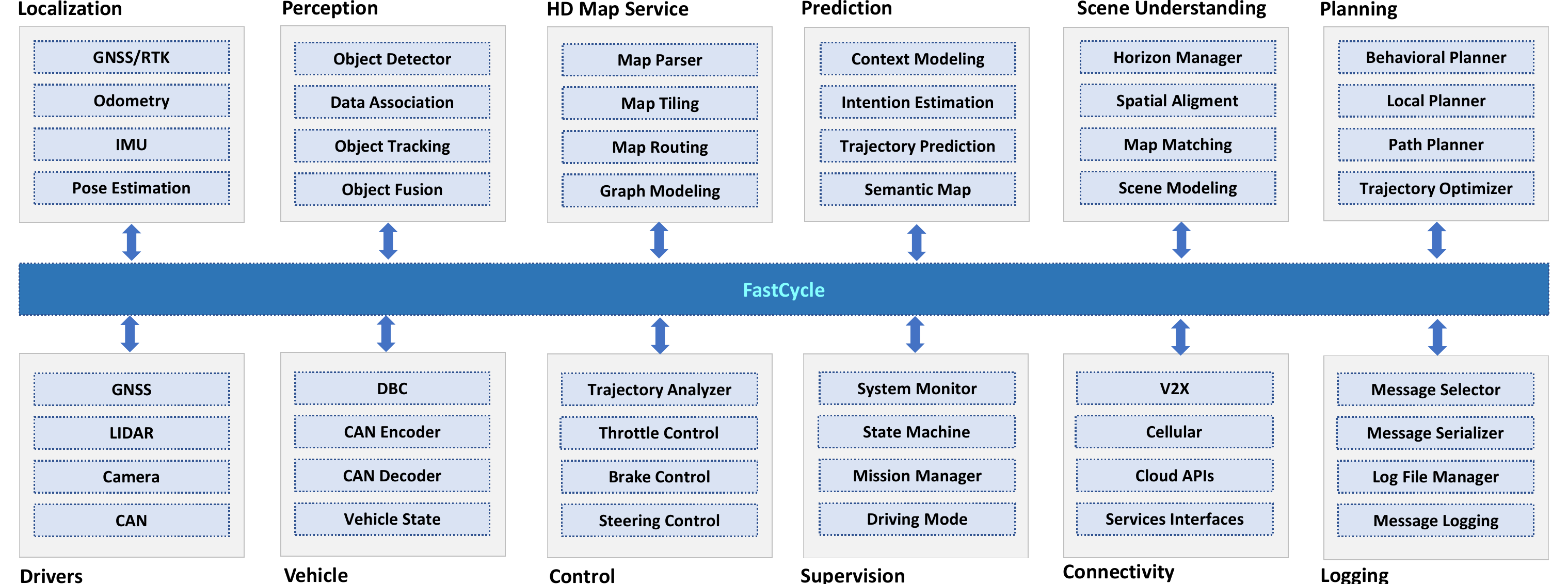}
\caption{High-level overview of the architecture of modern industry-level automated driving system}
\label{fig:had-architecture}
\end{figure*}

ADS development has witnessed considerable progress during the last few years \cite{yurtsever2020survey}. The ultimate goals of this technology are to provide a safe, comfortable and efficient mobility solution as well as minimizing road fatalities caused by human errors. Robustness of performance and reliability are both crucial if the potential of automated driving is to come to reality \cite{belluardo2021multi}. The key challenge in building ADS lies in the ability to meet stringent requirements, particularly of real-time performance and security. The ability to handle large amounts of sensor data, process them, make safe operational actions and share between the different subsystems, while rigorously respecting real-time constraints are critical in ADS \cite{wu2021oops}.
While tremendous efforts have been expended to improve data processing including the optimization of algorithms as well as hardware acceleration \cite{lin2018architectural}, transmission of data between the different components of the system is still a limiting performance bottleneck \cite{wu2021oops, liu2020robotic}. In this work, we present \textit{FastCycle}, a simplified middleware with efficient data transmission capabilities required for modular ADS.  

Several middlewares are already available for robotics and ADS development (ROS1\&2 \cite{ros}, CyberRT \cite{cyberrt}, TZC \cite{wang2019tzc}, Robust-Z \cite{liu2020robotic}, eProsima FastRTPS \cite{eProsima2022} and Nerve \cite{cruz2012dds}). Several automated driving software platforms have been developed based on these frameworks \cite{apollo, autoware}. 
Early releases of Baido Apollo and Autoware were based on ROS. Due to the performance limitations of ROS, Apollo developed their own middleware, CyberRT, with real-time and task-scheduling features, whereas Autoware migrated to ROS2. More details about the performance of both frameworks can be found in these works \cite{li2021autoware_perf, raju2019performance}. A middleware is a software layer enabling different communications, data management and task scheduling services between an operating system and an application. The middleware APIs provide the aforementioned services and motivate software developers to build their solutions on top of them. Middlewares are used to build not only distributed systems but also to build complex systems in a single machine. The architecture of a modular ADS, as shown in Fig. \ref{fig:had-architecture}, is composed of various subsystems that need to communicate and be managed by a middleware. Furthermore, middlewares enable developers to manage complex tasks more easily thanks to the abstraction of low-level details in a high-level API layer, which significantly reduces the development effort of an application. 

The communication mechanism of a middleware has a direct impact on the performance, reliability and security of the ADS to be developed \cite{wu2021oops}. Most of the existing middlewares in robotics and automated driving use Inter-Process Communication (IPC) over sockets as the backbone of their data transmission models \cite{basheer2019overview}. Middlewares built on top of sockets tend to be a highly reliable solution due to the loosely coupled communication mechanism offered by the TCP protocol \cite{liu2020robotic}. In other words, the crash of one process will not cause the whole system to crash. However, IPC over sockets always leads to high latency overhead and cannot guarantee good performance \cite{wu2021oops, liu2020robotic}. Conventional robotics middlewares such as ROS \cite{ros} utilise both TCP and UDP sockets for IPC. This allows processes to communicate either locally or between several machines. Nevertheless, messages are subjected to be copied several times resulting in unnecessary memory copies and system calls \cite{liu2020robotic}, explaining the high latency overhead of this mechanism. To improve socket-based IPC, Data Distribution Service (DDS) protocol \cite{dds, cruz2012dds} is proposed as a  reliable multicast solution over plain UDP sockets \cite{dds}. ROS2 \cite{ros} uses a fast implementation of DDS called FastDDS (formerly FastRTPS) developed by eProsima \cite{eProsima2022}. More details about the performance of ROS2 in comparison with ROS is given in \cite{maruyama2016exploring}. 

To overcome some of the drawbacks of socket-based IPC, shared memory techniques could be an alternative to reduce the memory copy operations. As the name implies, in a shared memory mechanism, two or more processes share the same memory space to pass messages, which has a significant impact on the performance of IPC. CyberRT is one of the high performance middlewares built using shared memory \cite{cyberrt}. The underlying mechanism is based on a shared buffer together with a ring structure divided into several slots. Each slot stores its state as well as message data. The state of a slot allows for only one process to access the data. Although shared memory based methods tend to be the most efficient implementation for IPC, serialization and deserialization of messages are still needed \cite{cyberrt}. As pointed in \cite{wu2021oops}, serialization and deserialization as well as data copy operations are the root cause of communication overhead latency. Furthermore, in IPC based middlewares, access to messages by third-party applications tends to be feasible which puts the security of these middlewares in question, mainly for automated driving applications. Moreover, an additional security layer would harm the overall performance \cite{kim2018security, rivera2019rosploit}. 

In intra-process based middlewares, communication between different threads (in the same process) offers a zero-copy data transport, which significantly improves performance. The gain in performance of this mechanism is obviously better in ROS nodelet, ROS2 and CyberRT intra-process communications, compared to their IPC counterparts \cite{wu2021oops}. However, messages in these middlewares are still exposed to external applications, e.g. for logging and visualization purposes \cite{ros, cyberrt}. Furthermore, implementations of a modular architecture based intra-process middlewares tend to be complicated. The present paper proposes \textit{FastCycle} as an efficient,  multi-threaded and modular intra-process based middleware for automated driving using zero-copy message transfer. \textit{FastCycle} is inherently secured since messages circulating between the different threads are not exposed to other external processes. Additionally, no message serialization and deserialization is needed except for logging. A comparative study based on standard metrics validates the performance of \textit{FastCycle}.

The rest of this paper is organized as follows. The high-level architecture of ADS as well as their design requirements in terms of modularity, performance and security are presented in Section \ref{sec:had-architecture}. The architecture and functionality of \textit{FastCycle} is presented in Section \ref{sec:fastcycle-architecture}. 
The evaluation of performance of \textit{FastCycle} compared with standard middlewares used in ADS is presented in Section \ref{sec:evaluation}. Finally, Section \ref{sec:conclusion} concludes this paper and highlights directions for future work. 

\section{AD System Architecture Requirements}
\label{sec:had-architecture}

Automated vehicles are sophisticated cyber-physical systems with a high-degree of interdependency and interoperability between their different components. The synergy between these components plays a crucial role in the robustness, reliability and the overall performance of an ADS. A key element in the design of its architecture is to define the mechanism these components use to share data between them. The message transfer mechanism has a substantial impact on the stability and performance of the overall system. In this section, we present a high-level architecture of a modular ADS. Based on this architecture, we elaborate on the design requirements of message sharing frameworks. 

\subsection{Automated Driving System Description}

Different ADS architectures exist in previous works.  
These architectures are mostly built on top of a middleware that manages the resources of the software components and transfer data between them. Fig. \ref{fig:had-architecture} shows a typical architecture of a modular ADS. The various components constituting the overall system need to communicate with one another in an efficient manner, to meet performance and security requirements. Popular middlewares such as ROS and CyberRT have been proposed as development frameworks for ADS \cite{hellmund2016robot, apollo}. Different architectures of ADS have been proposed in the literature \cite{tacs2016functional, ulbrich2017towards, reke2020self, lattarulo2021audric2}. Understanding the architecture and the different components of an ADS is essential to analyse the design requirements. Similar to robotics systems, an automated vehicle is considered as a cognitive agent. From an abstract point of view, it consists of sensing, cognitive and action elements. In more details, a modern industry-level ADS is typically composed of 12 modules as depicted in Fig. \ref{fig:had-architecture}.
The role of the \textit{drivers} module is to read and pre-process raw data from various sensors such as camera, lidar, IMU, GPS, sonar, radar, CAN bus and to provide them to other modules.
The \textit{localization} module provides the rest of the system with a precise estimation of the vehicle position, velocity and acceleration with respect to the environment.
The \textit{perception} module receives sensor data to detect obstacles (e.g. vehicles, pedestrians, and cyclists) as well as traffic signs, lanes and road borders. High definition (HD) digital maps are usually stored in remote servers. The \textit{HD map service} module is responsible for downloading a portion of this map (map tile) which is needed by various parts of the system to understand the environment and improve localization of the vehicle. Additionally, this module calculates the shortest route to the destination point. 
Then \textit{scene understanding} module receives an HD map tile as its input as well as vehicle localization, raw perception objects and semantic images. The role of this module is to understand what are the relevant objects for decision making and in what context they exist, e.g. a pedestrian is on a crossing or a sidewalk. 
The \textit{prediction} module receives as its input the vehicle localization, a list of relevant objects from the scene understanding module and it predicts their behavior by computing their predicted trajectories.
The \textit{motion planning} module generates a safe and collision-free reference driving trajectory for the automated vehicle. To do so, this module needs an HD map tile, a route to the destination as well as the predicted trajectories of relevant objects generated by the prediction module. The \textit{control} module receives the planned trajectory and computes control commands for the steering, brake and acceleration actuation systems. Since those actuators are available through vehicle CAN bus, it is the \textit{vehicle} module that is responsible for encoding/decoding raw CAN frames. One of the important modules for safety is the \textit{supervision} module. It defines the states and transitions to safely enter and exit the autonomous driving mode of the vehicle as well as to provide a visualization of the whole system. The \textit{connectivity} module enables communication with other vehicles as well as the smart infrastructure and the cloud. Finally, the \textit{logging} module records all (or selected) events/messages in the system for later use in incidents analysis and debugging.

\subsection{Design Requirements}

A crucial step in the architecture design of an ADS is the choice of the development framework on which the overall system will be developed. As discussed earlier in this section, an ADS is a very complex system from an architectural point of view. To ensure a robust behavior of the system, certain key design requirements are important. 

\subsubsection{Performance}

The overall performance of the system depends on the capability to transfer messages of typical payload such as raw images, point clouds, and map tiles from one subsystem to another as well as to process them in real-time. Modules involved in the control loop of the system, \textit{i.e.} from perception to action must be harmonized at a frequency depending on the physics of the system. The system should be able to complete all needed computations within a specific period of time in a deterministic way. 

\subsubsection{Modularity}

A development framework should allow to build a complex system that is made up of independent software modules that can be developed, maintained, tested and integrated with other modules with a minimum effort. The middleware used should allow to define common input and output interfaces as well as encapsulation of all metadata and configuration parameters to allow for further extensions.

\subsubsection{Security}

In a complex distributed system as an ADS, the communication between the various subsystems must be secured against unauthorized data access by external processes. 

\subsubsection{Reliability} 

The ability to implement a reliable and fail-safe ADS and capable of recovering from failures is crucial. Reliability in IPC-based middleware is feasible since the crash of one process does not necessarily lead the others to crash. However, in a multi-threaded architecture, one must rely on a robust implementation and maybe redundancy of the whole system.

\section{FastCycle Architecture}
\label{sec:fastcycle-architecture}

\textit{FastCycle} is a lightweight multi-threaded zero-copy message sharing framework. In this section we discuss its architecture in more details together with the various building blocks of the system, as well as the transfer model of the message sharin\textbf{}g mechanism.

\begin{figure}[!t]
\centering
\includegraphics[width=\linewidth]{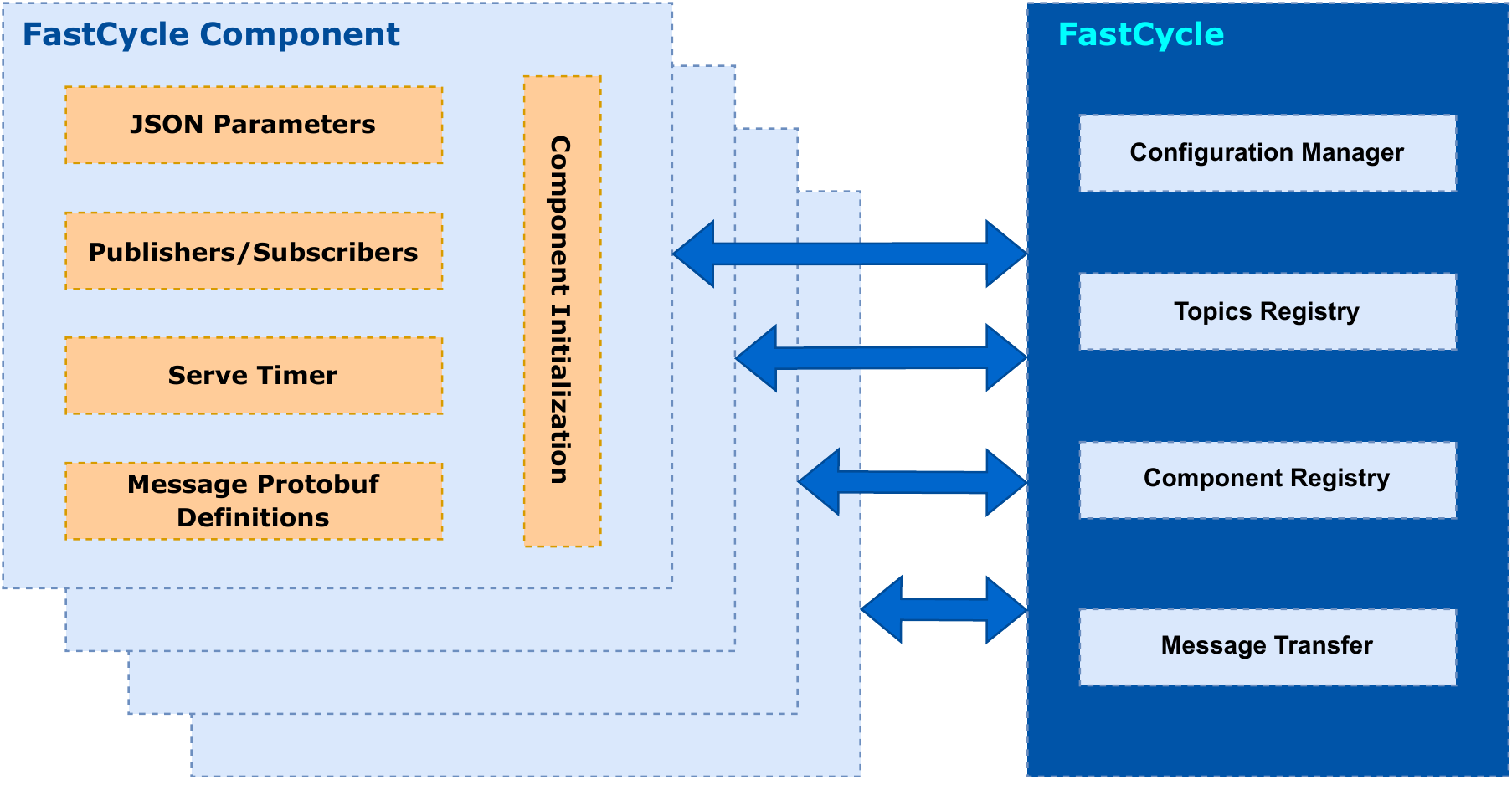}
\caption{FastCycle high-level architecture}
\label{fig:fastcycle-architecture}
\end{figure}

\subsection{FastCycle Anatomy}

The main element in building a modular ADS software is the \textit{FastCycle} component. It encapsulates everything needed to build a subsystem. As shown in Fig. \ref{fig:fastcycle-architecture}, a component defines the inputs and outputs, e.g. via \textit{subscribers}, \textit{publishers}, configuration parameters, and message definitions. The core system of \textit{FastCycle} is composed of the following five main elements:  

\subsubsection{Configuration Manager} It handles the parameters included in a configuration file associated with each component. The configuration file is written in JSON format and it defines all needed parameters that control the component behavior. For instance, if the component implements a control module of an ADS, this file will typically contain the sample period, parameters of the control laws for steering, brake, and acceleration, as well as the physical model parameters of the vehicle and its actuators. 

\subsubsection{Message broker}
The message broker is the core of \textit{FastCycle} and is responsible for the transfer and delivery of messages from a publisher to a corresponding subscriber. Its main components are a topics registry, a main thread, and a thread-pool. The topics registry keeps a record of the subscribers of a given topic. The main thread checks for newly published messages and submit the callback of a corresponding subscriber to the thread-pool.

\subsubsection{Message Transfer} This mechanism handles the zero-copy message transmission between the different components registered in \textit{FastCycle}. More details about the message transfer mechanism are given later in this section. 

\subsubsection{Component Registry} All components to be executed by \textit{FastCycle} are listed in a configuration file which contains the component names as well as a their parameters. At the most abstract level, each component is compiled as a shared library and if its name is included in the configuration file, they will be added to the component registry, which will then be used by the main thread of \textit{FastCycle} for execution. A component is derived from a base class giving access to all APIs needed for initialization with parameters and defining publishers and subscribers. A callback function is associated with each subscriber that is invoked once a new message is received. Each component has also access to one \textit{Serve Timer} called at a configurable cycle period.

\subsection{Message Transfer Model}
\begin{figure}[!t]
\centering
\includegraphics[width=\linewidth]{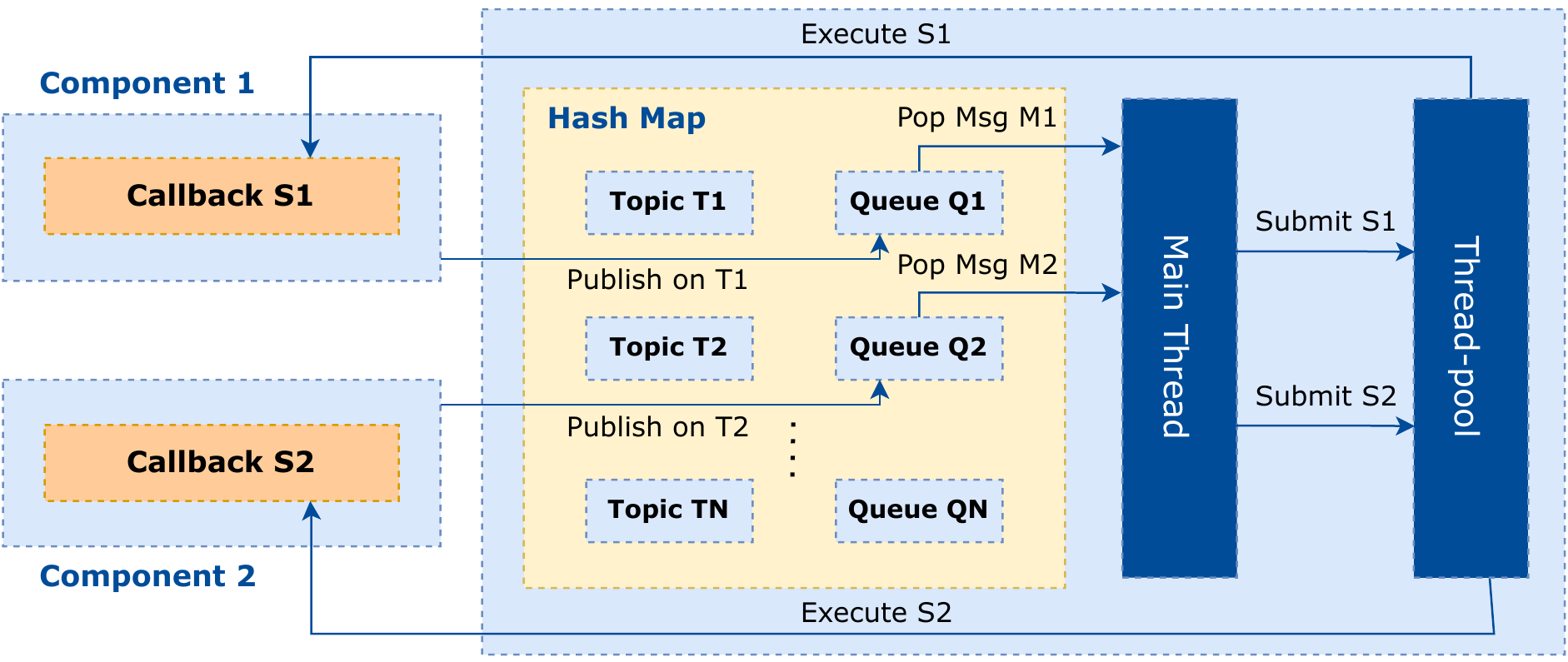}
\caption{FastCycle message transfer model}
\label{fig:fastcycle_core}
\end{figure}

The performance achieved by \textit{FastCycle} is due to its multi-threaded architecture and zero-copy message transfer between its components. 
The topics registry is implemented as a hash map whose keys are the registered topic names and the values are the message queues for each corresponding subscriber. Message queues are updated by each publisher while the main thread will check for newly published messages and submit the message callback of a corresponding subscriber to the thread-pool as depicted on Fig. \ref{fig:fastcycle_core}. Only a shared pointer is given to the subscriber callback, thus ensuring a zero-copy transfer. Moreover, in order to avoid race conditions, the content of the pointer cannot be modified.

\subsection{Message Format}

The message format used in \textit{FastCycle} is Google's protocol buffers (or simply \textit{protobuf}). Protobuf offers a small-size, simple and fast serializable/deserializable message representation \cite{protobuf2022}. Similar to ROS message format, protobuf supports the common primitive data types such as boolean, strings, integers and floating points. These primitives allow to define more customised message definitions. Furthermore, protobuf guarantees backward compatibility with earlier versions, which is an essential feature of extensibility. 
Protobuf compiler takes user-defined messages (.proto file) and generates compiled classes accessible in code. Due to these features, CyberRT has adopted protobuf as a message format since the beginning \cite{apollo, cyberrt} and so does \textit{FastCycle}. 

\section{Evaluation and Comparative Study}
\label{sec:evaluation}

In order to assess the performance of our proposed framework, we conduct a comparative study with well-known frameworks in the robotics and automated driving, namely ROS1, ROS2 and CyberRT. Both ROS1 and ROS2 will be evaluated in both inter-process and intra-process communication modes. 
In inter-process mode, messages are exchanged using either socket-based (ROS1) or DDS-based (ROS2) IPC mechanisms. On the other hand, in intra-process mode, messages are locally shared, meaning that publishers and subscribers spin inside the same process. ROS1\&2 intra-process are expected to be faster than their inter-process counterparts. This section compares the performance of these messaging frameworks, as well as provides a general assessment of their strengths and weaknesses for their uses in the context of autonomous driving.

\subsection{Methodology}

Our methodology to assess the performance of \textit{FastCycle} in comparison with other frameworks is based on three different metrics measured against different message payloads. The \textit{latency} is widely used to assess the time elapsed from the generation of a message in one component to its reception by another component in the system. We also use the \textit{Round-Trip Time} (RTT) which quantifies the time taken from sending a message to its reception back in the same component. The third metric we use is the \textit{jitter}, which quantifies the difference between the sending frequency and reception frequency thus assessing asymmetry in real-time capabilities between emitting and receiving components. These metrics have been used to evaluate the performance of different frameworks \cite{hellmund2016robot, profanter2019opc, gavrilov2021analysis, wu2021oops, maruyama2016exploring, kronauer2021latency}.
Let $i$ be the index of the $i$th message, the metrics we use are defined as follows.

The latency $l_i$ is defined as:
\begin{equation}
\label{eq:latency}
    l_i = T_{i,S} - T_{i,P},
\end{equation}
where $T_{i,S}$ is the reception time and $T_{i,P}$ is the publication time.

The RTT $r_i$ is defined as:
\begin{equation}
\label{eq:rtt}
    r_i = T_{i,P_1} - T_{i,P_0},
\end{equation}
where $T_{i,P_1}$ is the reception time of the echoed message $i$ and $T_{i,P_0}$ is the publication time.

The jitter $j$ is defined as the standard deviation of the latency:
\begin{equation}
\label{eq:jitter}
    j = \sigma(l_i),
\end{equation}
Our scenario for the comparative study is to measure latency, RTT and jitter as defined in equations (\ref{eq:latency}), (\ref{eq:rtt}) and (\ref{eq:jitter}) respectively at different message payloads. As discussed in Section \ref{sec:had-architecture}, a message sharing framework should be capable of handling messages of sizes typical to raw images, point clouds and map tiles. A reasonable payload of these messages would be up to 4MB \cite{wu2021oops}. For statistical relevancy, we measure our performance metrics over a relatively big number $n$ of samples ($n = 5000$ samples are used in these experiments). The sending interval for the messages is set to 1ms.

\subsection{Experimental Setup}

The experiments are conducted on an industrial PC which is used as an autonomous driving computer by our Junior test vehicle \cite{varisteas2019junior}. It is a Sintrones ABOX-5200G with an Intel Core i7-8700T CPU, 32GB RAM and one Nvidia GTX 1060 GPU.
The operating system used for all experiments is Ubuntu 20.04 with a kernel version 5.4. Docker containers with similar settings are used to setup the test environment for each framework. \textit{FastCycle} has been evaluated against ROS Noetic and Apollo CyberRT x86\_64-18.04.

\renewcommand{\arraystretch}{1.25}

\begin{table*}[!t]
    \centering
    \begin{tabular}{|p{1.6cm}||p{2.0cm}|p{2.0cm}|p{2.0cm}|p{2.0cm}|p{2.0cm}|}
    \hline
        & \textbf{32KB} & \textbf{128KB} & \textbf{512KB} & \textbf{1MB} & \textbf{4MB} \\
        \hline
        CyberRT & 2039 $\pm$ 37 & 3500 $\pm$ 42 & 14647 $\pm$ 65 & 29536 $\pm$ 86 & 119137 $\pm$ 385\\
        ROS1 - inter & 558 $\pm$ 25 & 650 $\pm$ 31 & 1305 $\pm$ 109 & 2017 $\pm$ 164 & 6196 $\pm$ 487 \\
        ROS2 - inter & 243 $\pm$ 10 & 299 $\pm$ 10 & 473 $\pm$ 8 & 908 $\pm$ 45 & 3602 $\pm$ 1271 \\
        ROS1 - intra &  208 $\pm$ 29 & 266 $\pm$ 26 & 468 $\pm$ 45 & 803 $\pm$ 50 & 1504 $\pm$ 40 \\
        ROS2 - intra & \textbf{146} $\pm$ 20 & 166 $\pm$ 22 & 237 $\pm$ 38 & 299 $\pm$ 3 & 796 $\pm$ 167 \\
        FastCycle & 159 $\pm$ 164 & \textbf{91} $\pm$ 98 & \textbf{218} $\pm$ 142 & \textbf{205} $\pm$ 133 & \textbf{289} $\pm$ 115\\
        \hline
    \end{tabular}
    \caption{Latency mean and standard deviation in $\mu$s for different message sizes}
    \label{tab:latency}
\end{table*}

\begin{table*}[!t]
    \centering
    \begin{tabular}{|p{1.6cm}||p{2.0cm}|p{2.0cm}|p{2.0cm}|p{2.0cm}|p{2.0cm}|}
    \hline
        & \textbf{32KB} & \textbf{128KB} & \textbf{512KB} & \textbf{1MB} & \textbf{4MB} \\
        \hline
        CyberRT & 3018 $\pm$ 706 & 6983 $\pm$ 71 & 29402 $\pm$ 624 & 59885 $\pm$ 1906 & 240008 $\pm$ 5119\\
        ROS1 - inter & 1019 $\pm$ 45 & 1350 $\pm$ 70 & 2326 $\pm$ 45 & 3629 $\pm$ 59 & 10658 $\pm$ 643 \\
        ROS2 - inter & 471 $\pm$ 49 & 535 $\pm$ 35 & 850 $\pm$ 28 & 1779 $\pm$ 88 & 2601 $\pm$ 253 \\
        ROS1 - intra &  238 $\pm$ 23 & 280 $\pm$ 32 & 492 $\pm$ 47 & 796 $\pm$ 53 & 2494 $\pm$ 323 \\
        ROS2 - intra & 214 $\pm$ 27 & 272 $\pm$ 42 & 448 $\pm$ 42 & 669 $\pm$ 33 & 1665 $\pm$ 71 \\
        FastCycle & \textbf{133} $\pm$ 134 & \textbf{162} $\pm$ 142 & \textbf{166} $\pm$ 136 & \textbf{277} $\pm$ 144 & \textbf{537} $\pm$ 78\\
        \hline
    \end{tabular}
    \caption{RTT mean and standard deviation in $\mu$s for different message sizes}
    \label{tab:rtt}
\end{table*}

\begin{table}[t]
\centering
\begin{tabular}{ 
|p{1.6cm}||p{1.2cm}|p{1.4cm}|p{1.2cm}|p{1.2cm}|}
 \hline
  & Modularity & Performance & Security & Reliability \\
 \hline 
 CyberRT & \centering ++ & \centering --  & \centering - &  ++ \\
 \hline
 ROS1 - inter & \centering  ++ & \centering -- & \centering - &  ++ \\
 \hline
 ROS2 - inter & \centering  ++ & \centering  - & \centering  - &  ++ \\
 \hline
 ROS1 - intra & \centering ++ & \centering + & \centering -  &  + \\
 \hline
 ROS2 - intra & \centering ++ & \centering ++ & \centering  - &  + \\
 \hline
 FastCycle  & \centering  ++ & \centering  ++ & \centering ++ &  + \\
 \hline
\end{tabular}
\caption{Conclusive comparison between different software frameworks}
\label{tab:comparison}
\end{table}

\subsection{Evaluation and Comparison}

TABLE \ref{tab:latency} and \ref{tab:rtt} show the results of the aforementioned evaluation scenarios. \textit{FastCycle} achieves better performance in terms of mean latency and RTT in almost all tests. 
The exception being a slightly greater mean latency than ROS2 intra-process for a message size of 32KB. 
Overall, ROS2 intra-process comes as the second fastest middleware thus demonstrating the improvements made over ROS1. It can also be seen that the effect of the message size is less sensitive in \textit{FastCycle}, whereas a steadier increase is observed in the other middlewares, especially in CyberRT and ROS1\&2 inter-process. \textit{FastCycle} is also 2 to 3 times faster than ROS2 intra-process for the 4MB message size, which demonstrates its substantial capabilities for large payloads. 
As for the jitter, \textit{FastCycle} seems to be under-performing as the standard deviations of the latency is mostly above other frameworks, although it catches up at 4MB. 
The higher values of jitter might be the result of some thread-pool induced delays and should be investigated in future work. Consequently, ROS2 intra-process seems to be more reliable under 4MB from a real-time perspective although it is slower on average. 
At 4MB however, \textit{FastCycle} has the smallest standard deviation which confirms the better performance at higher message payloads. It can also be worth noting that the CyberRT performance across the board seems exceptionally poor.

As a conclusive comparison, CyberRT and ROS1\&2 inter-process are modular, extensible and reliable frameworks thanks to their multi-process designs, however, this in turn induces a significant performance overhead. Furthermore, the messages are exposed to other processes at the operating system level which can lead to security vulnerabilities. ROS1\&2 intra-process are very similar but offer a remarkable performance improvement at the cost of reliability. Across the board, ROS2 is better than ROS1, especially at higher payloads. Finally, \textit{FastCycle} showed the best latency and RTT performance although it falls behind ROS2 in jitter. \textit{FastCycle} was particularly fast at higher payloads which is important for autonomous driving applications. \textit{FastCycle} is also modular and safe as it does not expose messages to external applications, however, its reliability falls behind the inter-process counterparts. The above discussion is summarized in Table \ref{tab:comparison}.

\section{Conclusion and Future Work}
\label{sec:conclusion}

This paper presented \textit{FastCycle} as a lightweight multi-threaded zero-copy message sharing framework that has been designed to meet the requirements of a high fidelity ADS in terms of modularity, performance and security. The architecture of \textit{FastCycle} as well as its different components along with the message transfer model and format are presented. Evaluation of \textit{FastCycle} based on latency and RTT in comparison with ROS and CyberRT shows the improved performance of the proposed framework although future work should improve upon the current jitter and subsequently real-time capabilities. Based on the evaluation study, \textit{FastCycle} is chosen as the middleware in our modular autonomous system \textit{RoboCar}, which is currently under development. The current version of \textit{FastCycle} still uses \textit{catkin} package management system, which is also left for future work. 

\section{Acknowledgment}

This work has been financially supported by the EU INTERREG GR Terminal project.

\bibliographystyle{ieeetr}
\bibliography{references}

\end{document}